\documentclass[10pt,conference,a4paper]{IEEEtran}
\usepackage{graphicx}
\usepackage{multirow}
\usepackage{balance}
\usepackage[utf8x]{inputenc}
\usepackage[table,xcdraw]{xcolor}

\ifCLASSINFOpdf
\else
\fi
\begin{document}
\title{Viability of Optical Coherence Tomography for Iris Presentation Attack Detection}

\author{\IEEEauthorblockN{Renu Sharma, Arun Ross}
\IEEEauthorblockA{Michigan State University\\
Email: \{sharma90,rossarun\}@cse.msu.edu}
%\and
%\IEEEauthorblockN{Arun Ross}
%\IEEEauthorblockA{Michigan State University\\
%East Lansing, MI 48824, U.S.\\
%Email: rossarun@cse.msu.edu}
}


%Telephone: (800) 555--1212, Fax: (888) 555--1212}


\maketitle

\begin{abstract}
 %In this paper, we report the results of a study, where the efficacy of multiple imaging modalities is evaluated for the problem of iris presentation attack (PA) detection. The 3 modalities considered are near-infrared imaging (NIR), visible spectrum imaging (VIS) and optical coherence tomography imaging (OCT). 
In this paper, we propose the use of Optical Coherence Tomography (OCT) imaging for the problem of iris presentation attack (PA) detection. We assess its viability by comparing its performance with respect to traditional iris imaging modalities, viz., near-infrared (NIR) and visible spectrum. OCT imaging provides a cross-sectional view of an eye, whereas traditional imaging provides 2D iris textural information. 
%In this paper, we introduce the use of Optical Coherence Tomography (OCT) imaging for the problem of iris presentation attack (PA) detection. OCT imaging provides cross-sectional view of an eye. Existing literature utilizes near-infrared (NIR) and visible spectrum imaging, which provides 2D iris textural information.
% This is the first time an OCT-based approach is presented in the context of iris PA detection.   
%Viability of OCT in iris PA detection is also assessed by comparing it against near-infrared (NIR) and visible spectrum imaging. 
PA detection is performed using three state-of-the-art deep architectures (VGG19, ResNet50 and DenseNet121) to differentiate between bonafide and PA samples for each of the three imaging modalities. Experiments are performed on a dataset of 2,169 bonafide, 177 Van Dyke eyes and 360 cosmetic contact images acquired using all three imaging modalities under intra-attack (known PAs) and cross-attack (unknown PAs) scenarios. We observe promising results demonstrating OCT as a viable solution for iris presentation attack detection.
\end{abstract}


% \fi
\IEEEpeerreviewmaketitle

\section{Introduction}
An iris recognition system recognizes an individual based on their iris texture. A majority of commercial iris recognition systems acquire iris images in the near-infrared (NIR) spectral band. Recently, iris recognition systems have been deployed on mobile devices that capture the iris in the visible spectrum (VIS).

Iris systems are vulnerable to presentation attacks (PAs) where an adversary presents an artificial or altered biometric sample to the sensor to obfuscate their own identity or to spoof someone else's identity or to create a virtual identity \cite{Ross2019}. Examples of PAs include printed eye, artificial eye (artificial eye or plastic eye), replay attack, cosmetic contact lens, etc. Our objective is to detect iris PAs when presented to an iris recognition system. 

%Optical Coherence Tomography (OCT) imaging is also introduced for biometric imaging, but especially for fingerprint modality \cite{Bossen2010}, it is not yet used for iris recognition. Most of the iris recognition systems facilitate the unattended acquisition of iris image to automate the process. This setup also makes it vulnerable to presentation attacks. Presentation Attack (PA) occurs when
%Though, we have only considered artificial eyes and cosmetic contact contact lens in this work, our focus is to build a solution which is generalizable across unknown PAs. 

Techniques proposed in the literature to counter iris PAs can be categorized as software-based or hardware-based. Software-based techniques utilize the iris {\em image} captured by the standard iris sensor to detect PAs. Various texture-based features are extracted from the captured iris image to detect the PA, for instance, SIFT \cite{Zhang2015}, LBP \cite{Gragnaniello2015b} and BSIF \cite{Raghavendra2015}. More recently, deep-learning based approaches have been utilized for PA detection from the captured iris image \cite{Menotti2015,Chen2018a,Hoffman2019,Sharma2020}. The techniques referenced above are mostly based on NIR iris images except \cite{Gragnaniello2015b, Raghavendra2015} which operate on VIS images. Menotti \textit{et al.} \cite{Menotti2015} showed results on both NIR and VIS iris images. Raghavendra and Busch \cite{Ramachandra2014} exploited characteristics of the Light Field Camera (LFC) for iris PA detection in the VIS spectrum. Sequeira \textit{et al.} \cite{Sequeira2016} suggested the use of a one-class classifier on VIS images for generalization across unseen attacks, i.e., attacks that were not used in the training phase.  
%Yadav \textit{et al.} \cite{Yadav2017} provided baseline algorithm and dataset (UVCLI dataset) of cosmetic contact contact lens images in visible spectrum. 
In \cite{Raja2015}, the authors utilized Eulerian Video Magnification (EVM) to detect PAs in VIS videos. 

Hardware-based techniques require {\em additional hardware} in addition to the iris sensor to capture the liveness characteristics of the iris. Techniques include analyzing pupil dynamics \cite{Czajka2019}, 3D structural modeling of the eye using stereo imaging \cite{Hughes2013}, observing change in pupil size and iris texture with changes in light intensity \cite{Kanematsu2007} and gleaning eye movement cues \cite{Komogortsev2013}. There are some techniques that employ multi-spectral illumination for iris PA detection. Park and Kang \cite{Park2007} utilized a specialized tunable filter to capture iris images at different spectral bands ranging from 650nm to 1100nm. These multi-spectral images are then fused at the image-level to detect PAs. Lee \textit{et al.} \cite{Lee2006} analyzed the reflectance properties of the iris and sclera in multi-spectral illumination. Chen \textit{et al.} \cite{Chen2012} captured images at the near-infrared (860nm) and blue (480nm) wavelengths, and then analyzed the conjunctival vasculature patterns and the iris textural patterns for liveness detection.\footnote{Early literature used the term ``liveness detection" to refer to the problem of PA detection.} Connell \textit{et al.} \cite{Connell2013} exploited the anatomy and geometry of the human eye using structured light to detect cosmetic contact lens. Thavalengal \textit{et al.} \cite{Shejin2016} used both VIS and NIR images for iris liveness detection in smartphones. Hsieh \textit{et al.} \cite{Hsieh2018} utilized dual-band imaging hardware (VIS and NIR) to distinguish between the textured pattern of contact lens from real iris patterns using independent component analysis. 
% Recently, OCT imaging technique is also used in fingerprint recognition \cite{Bossen2010, Moolla2015}, fingerprint PA detection \cite{Moolla2019}, and  iris PA detection []. 
Assessment reports of other state-of-the-art iris PA detection techniques are presented in \cite{LivDet2017,Czajka2018,LivDet2020}. 

%As far as the OCT imaging is concerned, it has been reported to be used for fingerprint PA detection \cite{Moolla2019}. But, in the context of iris images, traditionally NIR and VIS imaging are used. NIR and VIS images capture the stromal textural patterns of the iris, whereas OCT images capture the internal structure of the eye and the iris (Figure \ref{fig:OCT_imaging}). We hypothesize that the use of OCT may aid in iris PA detection. But, the unavailability of OCT iris dataset and high hardware costs associated with OCT prevent its exploration in the field of iris PA detection. Currently, development of cost effective OCT hardware solutions is gaining momentum \cite{Song2019,Unterhuber2003,Nitkowski2014}. We anticipate that this will eventually result in an increasing trend in the study of OCT-based iris PAD. So, in an effort towards this, we are introducing OCT for iris PA detection for the first time to the best of our knowledge.

\begin{figure}[h]
	\centering
	\includegraphics[width=0.9\linewidth]{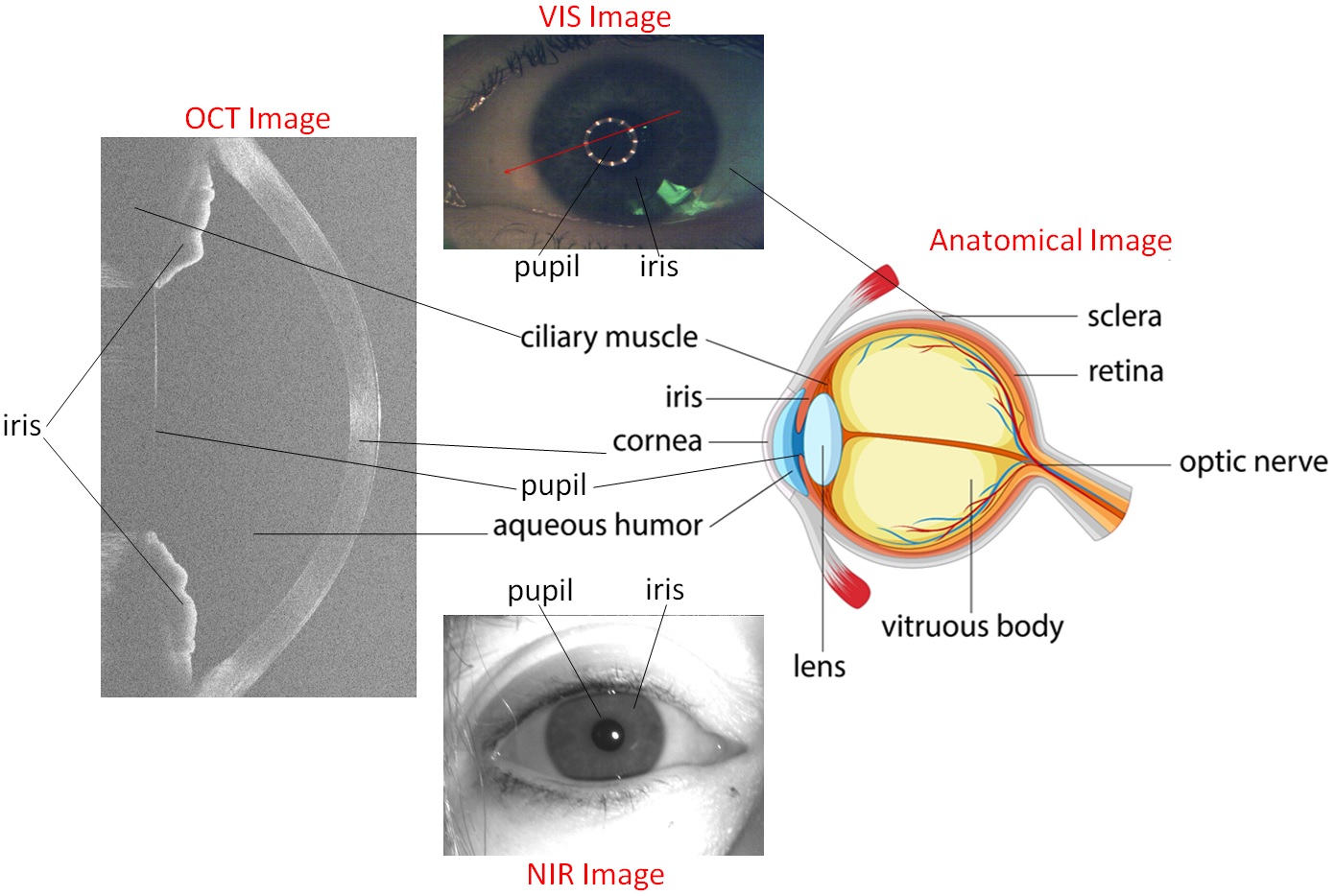}
	\caption{Components of the eye and iris sensed using OCT, NIR and VIS imaging. The anatomical image (\textit{https://www.vecteezy.com/vector-art/431288-parts-of-human-eye-with-name}) is also shown. The red line in the VIS image shows the traverse scanning direction of the OCT scanner.}
	\label{fig:OCT_imaging}
\end{figure}

In contrast to conventional PA detection algorithms based on NIR or VIS imaging, we propose a novel approach that uses Optical Coherence Tomography (OCT) imaging for iris PA detection. \footnote{OCT also employs NIR illumination, but obtains cross-sectional views not textural details.} NIR and VIS images capture the stromal textural patterns of the iris, whereas OCT images capture the internal structure of the eye and the iris (Figure \ref{fig:OCT_imaging}). OCT imaging has been utilized for fingerprint PA detection \cite{Moolla2019}. But the unavailability of an OCT iris dataset and the high hardware costs associated with OCT have traditionally prevented its exploration for iris PA detection. However, the development of cost-effective OCT hardware \cite{Song2019} motivates us to consider it for iris PA detection. To the best of our knowledge, this is the first time an OCT-based solution has been introduced for iris PA detection. The main contributions of our work are as follows:
\begin{enumerate}
	\item We propose a hardware-based iris PA detection technique based on OCT imaging technology. We also assess its viability by comparing its performance against traditional NIR and VIS imaging modalities.
	%We propose a novel approach of utilizing OCT imaging for iris PA detection. We also assess its performance by comparing it against NIR and VIS imaging modalities. 
	\item We implement OCT-based iris PA detection using three state-of-the-art deep CNN models which significantly differ in their architectures: VGG19 \cite{Simonyan2015}, ResNet50 \cite{He2015} and DenseNet121 \cite{Huang2018}.
	\item We evaluate PA detection performance on a dataset of 2,169 bonafide, 177 Van Dyke eyes and 360 cosmetic contact lens images under intra-attack and cross-attack scenarios. Each input sample is captured in all three imaging modalities. 
	\item We also generate CNN visualizations (heatmaps \cite{Selvaraju2017} and t-SNE plots \cite{Maaten2008}) to further analyze the results on OCT, NIR and VIS images. Heatmaps are used to identify salient image regions that the deep architectures utilize to detect PAs. t-SNE plots aid in visualization of features extracted by the CNN architectures.
\end{enumerate}

The rest of the paper is organized as follows. Section 2 discusses background of the imaging modalities. Section 3 describes the proposed approach. Section 4 provides a description of the dataset. Section 5 describes the experimental setup and reports the results. Section 6 concludes the work.

%Optical Coherence Tomography (OCT) is a non-invasive imaging technique used to capture 2D or 3D images of a sample (e.g., biological tissues). It uses long-wavelength light to penetrate deep into the sample and the light backscattered from the different layers of the sample form an image. The images created using OCT are of micrometer-resolution. 2D or 3D images of an eye captured using OCT imaging are mainly used for medical purposes (optometry, ophthalmology). Now, its use is further extended for detecting presentation attacks (PAs) in the iris modality. 2D cross-sectional image of iris taken at 1300nm wavelength provided structural information about the cornea, iris, pupil, lens as shown in Figure 1.

%-------------------------------------------------------------------------
\section{Background of Iris Imaging Modalities}
\begin{figure}[h!]
	\begin{center}
		\includegraphics[width=0.9\linewidth]{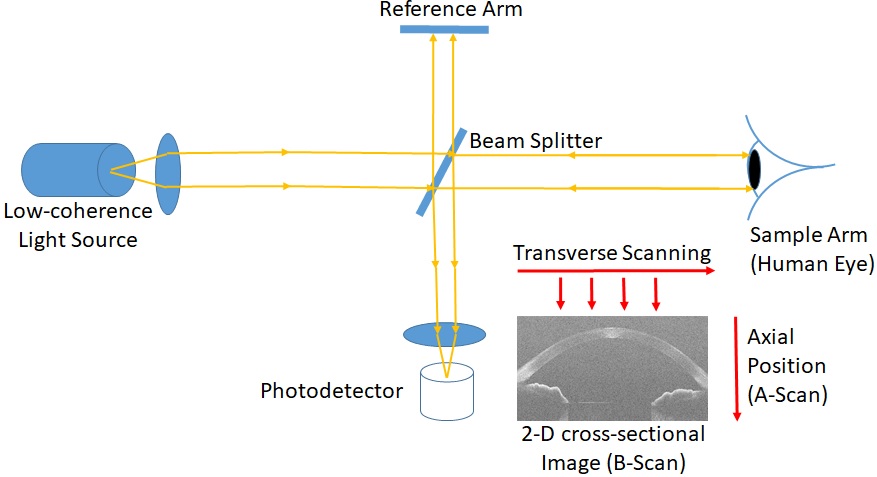}
	\end{center}
	\caption{Typical optical setup of an OCT scanner. Low-coherence light is incident over the beam splitter, which splits the light into sample and reference arms. Back-reflected light from sample and reference arms are then collected by the photodetector. Cross-sectional OCT image (B-scan) is formed by combining a number of A-scans along the transverse direction.}
	\label{fig:OCT-Sensor}
\end{figure}

The complex texture of the iris is characterized by its components, including, pigments (chromophore), blood vessels, muscles, crypts, contractile furrows, freckles, collarette and pupillary frills. Different spectral bands can potentially be used to capture different components of the iris. NIR illumination, which operates in the 700-900nm range, predominantly captures the stromal features (fibrovascular layer) of the iris, whereas VIS (400-700nm) captures information about the pigment melanin. 
%NIR spectrum eliminates pigment information due to the eumelanin component of chromophore which is not stimulated by NIR illumination.
Optical Coherence Tomography (OCT) \cite{Huang1991} is a non-invasive, micrometer-resolution imaging modality, that can be used to capture 2-D cross-sectional or 3-D volumetric images of an eye. It is mainly used for biomedical and clinical purposes, such as ophthalmology, optometry, cardiology and dermatology. It works with a low-coherence near-infrared (800nm-1325nm) light source. OCT imaging captures cornea (circular arc), iris tissue structure, anterior humor (the space between iris and cornea) and the ciliary muscles (next to the iris tissues) of the eye as shown in Figure \ref{fig:OCT_imaging}. OCT images are captured by shining the light source over a beam splitter, which splits the light into two beams, one directed to the sample arm (human eye) and another to the reference arm (mirror). The time delay and intensity of the back-reflected light from both the arms are estimated to create an axial back-scattering profile called A-Scan. Combination of A-Scans along transverse axis forms a 2-D cross-sectional image called B-Scan. The imaging setup of an OCT sensor is shown in Figure \ref{fig:OCT-Sensor}. OCT imaging primarily captures the structure and morphology of the eye as opposed to texture information that is typically observed in NIR and VIS images.  

A majority of commercial iris recognition systems and iris PA detection algorithms utilize NIR images for the following reasons. Firstly, NIR illumination penetrates deeper into the iris and elicits the textural pattern of both light and dark irides; in contrast, majority of VIS illumination is absorbed by higher levels of melanin in dark-colored irides resulting in poorly discernible iris texture. Secondly, background illumination variations and corneal reflections do not affect NIR imaging as much as RGB imagers. However, some iris recognition and PA detection algorithms have started using VIS imaging due to inexpensive hardware and a wide range of applications (mobile devices, surveillance, etc.) \cite{Gragnaniello2015b, Vyas2019}. Due to expensive hardware, OCT imaging has not been traditionally discussed in the literature for either iris recognition or PA detection. 
%Since it requires an additional sensor (OCT sensor) to capture the 2D cross-sectional image of an eye, it is categorized as a hardware-based technique. 
%Due to their popularity and efficiency, we have chosen these three imaging modalities for comparison.

\section{Proposed Approach}
\begin{figure}[h!]
	\begin{center}
		\includegraphics[width=\linewidth]{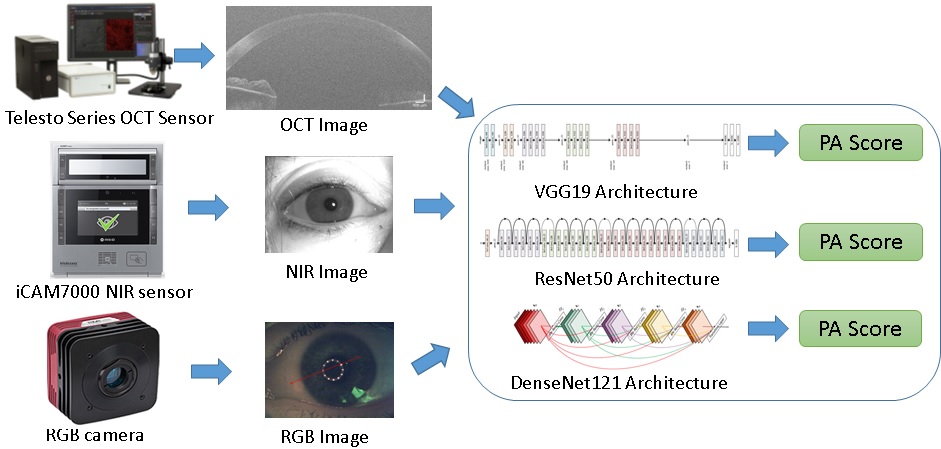}
	\end{center}
	\caption{Comparative analysis of OCT, NIR and VIS imaging in detecting iris PAs. Three architectures, viz., VGG19, ResNet50, DenseNet121, are used for distinguishing between bonafides and PAs by emitting a PA score. A higher PA score indicates the input is a ``PA" and a lower score indicates the input is a ``bonafide" image. OCT and RGB camera images are taken from \textit{https://www.thorlabs.com} and NIR sensor image is taken from \textit{https://www.irisid.com/}.}
	\label{fig:Proposed-Method}
\end{figure}

In this work, we discuss the use of OCT imaging for iris PA detection. For classification of iris OCT images as bonafide or PA, we used three state-of-the-art deep CNN architectures: VGG19 \cite{Simonyan2015}, ResNet50 \cite{He2015} and DenseNet121 \cite{Huang2018}. These architectures output a single PA score in the range [0, 1], with a `1' indicating a PA and `0' indicating a bonafide. Using the same CNN architectures, we compare the PA detection capability of OCT images against NIR and VIS images. Overview of the approach is depicted in Figure \ref{fig:Proposed-Method}. In the subsequent sub-section, we provide implementation details of all three network architectures.

To classify bonafide and PA iris images acquired from all three imaging modalities, we used three state-of-the-art deep architectures: VGG19 \cite{Simonyan2015}, ResNet50 \cite{He2015} and DenseNet121 \cite{Huang2018}. These three networks differ by the number of the convolutional layers, the number of trainable parameters and the connection type. VGG19 \cite{Simonyan2015} has 19 convolutional layers with kernels of fixed size 3 $\times$ 3 throughout the network. It has 143,667,240 trainable parameters. ResNet50 \cite{He2015} has 50 convolutional layers with residual connections (skip connections) to moderate gradient flow and allow the training of a large network. It has 35,610,216 trainable parameters. DenseNet121 \cite{Huang2018} consists of 121 convolutional layers, where each layer is connected to every other layer resulting in a much reduced set of trainable parameters (7,978,856). Three different sized architectures are utilized in the study to eliminate the bias created due to the network architecture (under-fitting or over-fitting) in the comparison results. As the dataset used in the study is insufficient to train these deep architectures, we utilize pre-trained models on ImageNet dataset. Pre-trained models also help in faster convergence during the training process. ImageNet is a large dataset used for object classification containing 1.2 million images of 1000 classes. The images in ImageNet dataset are visible spectrum images, i.e., RGB. To preserve the usefulness of pre-trained weights for the OCT and NIR spectrum images, we normalize OCT, NIR and VIS images using the mean and the standard deviation calculated from the ImageNet dataset images. The photometrically normalized images are then re-sized to 224 $ \times $ 224 and input to the aforementioned architectures. All three models are then fine-tuned using OCT, NIR and VIS iris images resulting in nine trained models. The learning rate used in the training is 0.005, the batch size is 20, the optimization algorithm is stochastic gradient descent with momentum of 0.9, the number of epochs is 50, and the loss function is cross-entropy. During test and evaluation, each of these networks produce a single PA score which is used along with a threshold to determine if the input image is a PA or a bonafide.

\section{Dataset}\label{dataset}
\begin{figure}
	\centering
	\includegraphics[width= 0.8\columnwidth]{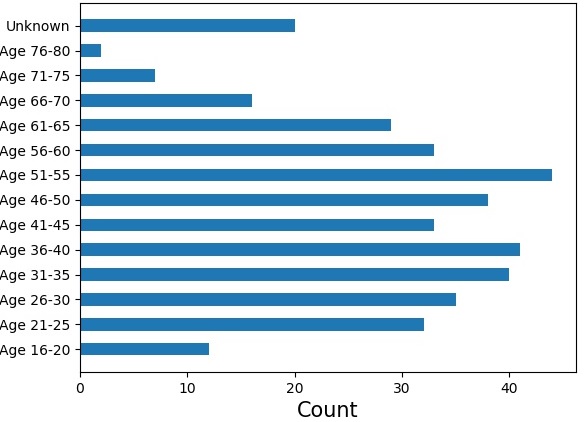}
	\caption{Age distribution of subjects in the dataset.}
	\label{fig:Age-Distribution}
\end{figure}

\begin{figure}
	\centering
	\includegraphics[width=\columnwidth]{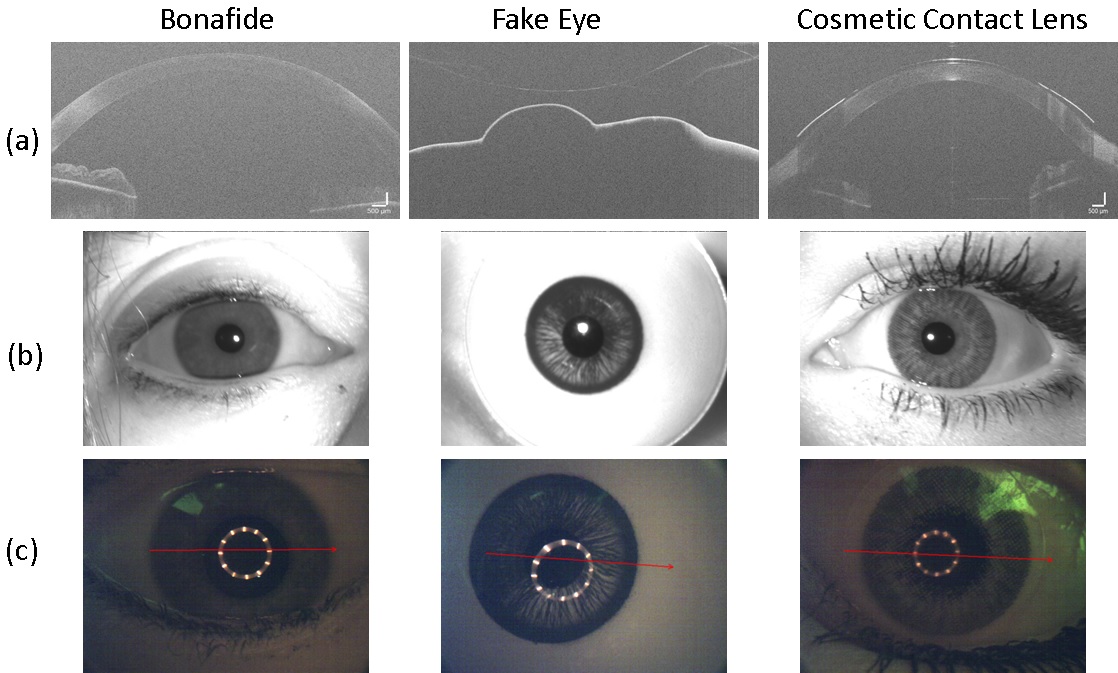}
	\caption{Samples of bonafide, artificial eyes and cosmetic contact lens images captured using (a) OCT, (b) NIR and (c) VIS imaging modalities.}
	\label{fig:Dataset-Samples}
\end{figure}

The dataset is collected under the Odin program of IARPA \cite{Odin} from 740 eyes (370 subjects). Figure \ref{fig:Age-Distribution} provides age distribution of subjects. The number of male and female subjects are 136 and 243, respectively. 
OCT, NIR and VIS images are collected sequentially for a subject using an RGB camera, iCAM7000 NIR sensor and THORLabs Telesto series (TEL1325LV2) OCT sensor \cite{OCTsensor}, respectively. The OCT images are acquired at 1325nm wavelength having 7mm imaging depth and 12$\mu$m axial imaging resolution. For a single sample, 50 cross-sectional frames are captured by the OCT sensor. However, temporal information is not significant among frames, so we use only the first frame. Iris PAs considered in this study are artificial eyes (Van Dyke eyes) and cosmetic contact lenses. For OCT and VIS, the dataset contains 844 bonafide images, 61 artificial eyes and 120 cosmetic contact lens images, whereas, for NIR, there are 1,371 bonafide images, 111 artificial eyes and 120 cosmetic contact lens images. Further sub-categorization of PA images is provided in Table \ref{table:Dataset_Details}. Figure \ref{fig:Dataset-Samples} shows examples of bonafide and PA images acquired in all three spectra (OCT, NIR and VIS).

\begin{table}[]	
	\centering
	\caption{Number of bonafide and PA samples corresponding to each imaging modality.}
	\label{table:Dataset_Details}
	\resizebox{0.95\columnwidth}{!}{%
		\begin{tabular}{|l|l|l|l|l|}
			\hline
			\multirow{2}{*}{\textbf{Classes}} & \multirow{2}{*}{\textbf{Sub-Classes}} & \multicolumn{3}{l|}{\textbf{Imaging Modality}} \\ \cline{3-5} 
			&  & \textbf{OCT} & \textbf{RGB} & \textbf{NIR} \\ \hline
			\multicolumn{2}{|l|}{Bonafide} & 844 & 844 & 1371 \\ \hline
			\multirow{3}{*}{Artificial Eyes} & Van Dyke Eye (Brown) & 30 & 30 & 51 \\ \cline{2-5} 
			& Van Dyke Eye (Blue) & 29 & 29 & 56 \\ \cline{2-5} 
			& Face Mask & 2 & 2 & 4 \\ \hline
			\multirow{3}{*}{Cosmetic Contacts} & Acuvue Accent Vivid & 37 & 37 & 43 \\ \cline{2-5} 
			& Air Optix Sterling Grey & 41 & 41 & 43 \\ \cline{2-5} 
			& Extreme FXS Halloween Blackout & 42 & 42 & 34 \\ \hline
		\end{tabular}
	}
\end{table}

\section{Experimental Setup and Results}

Before evaluating the three imaging modalities (OCT, NIR and VIS), we assess the performance of three fine-tuned architectures (VGG19, ResNet50 and DenseNet121) on the LivDet-iris 2017 \cite{LivDet2017} dataset for iris PA detection. The dataset is an amalgamation of Clarkson, Warsaw, Notre Dame and IIITD-WVU datasets. Print and cosmetic contact lens PAs are included in the dataset. The experimental setup is kept the same as specified in the competition \cite{LivDet2017}. Evaluation measures are Attack Presentation Classification Error Rate (APCER) and Bonafide Presentation Classification Error Rate (BPCER), where APCER is the proportion of PA samples misclassified as bonafide and BPCER is the proportion of bonafide samples misclassified as PAs. All three architectures either outperform or are comparable to the state-of-the-art algorithms (CASIA, Anon1 and UNINA) on the LivDet-iris 2017 competition as shown in Table \ref{table:LiveDet-2017-Results}. 

\begin{table}[]	
	\centering
	\caption{APCER (\%) and BPCER (\%) of all algorithms on LivDet-Iris 2017 Dataset \cite{LivDet2017}. Results are presented by averaging APCER and BPCER of all test sets in the dataset.}
	\label{table:LiveDet-2017-Results}
	\resizebox{0.95\columnwidth}{!}{%
		\begin{tabular}{|l|l|l|l||c|c|c|}
			\hline
			\textbf{Algo.} & \textbf{CASIA{\cite{LivDet2017}}} & \textbf{Anon1{\cite{LivDet2017}}} & \textbf{UNINA\cite{LivDet2017}} & \multicolumn{1}{l|}{\textbf{VGG19}} & \multicolumn{1}{l|}{\textbf{ResNet50}} & \multicolumn{1}{l|}{\textbf{DenseNet121}} \\ \hline
			APCER & 11.88 & 14.71 & 15.52 & 15.80 & 11.71 & \textbf{6.25} \\ \hline
			BPCER & 9.48 & 3.36 & 12.92 & \textbf{1.20} & 3.24 & 10.39 \\ \hline
		\end{tabular}
	}
\end{table}

Utilizing the three architectures, we perform comparative evaluation of OCT, NIR and VIS images in detecting iris PAs. Experiments are performed under intra- and cross-attack scenarios. Samples that were successfully captured in all three imaging modalities are selected for experiments. The dataset used for evaluation eventually has 723 bonafide samples, 59 artificial eyes and 120 cosmetic contact lens images captured in all three imaging modalities. The train, validation and test sets are {\em eye-disjoint}, i.e., they have data from different eyes and samples in the three sets are {\em mutually exclusive}. Intra-attack experiments examine which imaging modality performs best with known PAs (used during training), whereas cross-attack experiments analyze the generalizability across unknown PAs (not used in training). The evaluation measures used are True Detection Rate (TDR) at 0.2\% False Detection Rate (FDR), and Average Classification Error Rate (ACER). TDR is the percentage of PA samples that were correctly detected, whereas FDR is a percentage of bonafide samples that were misclassified as PA. ACER is the average of APCER and BPCER. Receiver operating characteristic (ROC) curves are also provided for a comprehensive overview. For successful detection, TDR should be comparatively higher and ACER should be comparatively lower.

\subsection{Intra-attack Setup and Results}
In the intra-attack setup, three experiments are performed: Intra-EXP 1, Intra-EXP 2 and Intra-EXP 3. Intra-EXP 1 includes both the PAs (artificial eyes and cosmetic contact lens) and bonafide images in the training and test sets, whereas Intra-EXP 2 and Intra-EXP 3 include images from only one PA along with bonafide images for training and testing. Intra-EXP 2 and Intra-EXP 3 experiments are performed to test the difficulty level of differentiating a specific PA from bonafide samples. Details about the train, validation and test sets of all three experimental setups are provided in Table \ref{table:Experimental-Setup}. In the first experiment (Intra-EXP 1), the data are split in a 70:30 ratio, where 70\% of eyes is used for training and the remaining for testing (30\%). Thereafter, five-fold cross-validation is employed on the training set, where 4 folds are used for training and one for validation. The validation set is used to estimate the threshold to be used on the test set for calculating ACER. The TDR at 0.2\% FDR and the ACER for VGG19, ResNet50 and DenseNet121 architectures are provided in Table \ref{table:All-Results}. ROC curves of Intra-EXP 1 for all three architectures are shown in Figures \ref{fig:VGG19-Results}(a), \ref{fig:ResNet50-Results}(a) and \ref{fig:DesNet121-Results}(a).

In the Intra-EXP 1 experiment, the best results are observed on OCT images, second-best on NIR images, and then on VIS images. All trained models (five) obtained from cross-validation show low standard deviation in the results when tested on OCT images (Figures \ref{fig:ResNet50-Results}(a) and \ref{fig:DesNet121-Results}(a)) compared to NIR and VIS images. Similar results are observed across all three network architectures (VGG19, ResNet50 and DenseNet121). This validates the robustness of PA detection when using OCT images. Considering individual PAs in Intra-EXP 2 and Intra-EXP 3 experiments, it is found that both types of PAs are perfectly classified (100\% TDR) by the OCT and NIR modalities. There are a few errors when detecting cosmetic contact PAs using the VIS modality (98.63\% TDR). So, in the intra-attack scenario, where attacks are known and used during training, the OCT modality perfectly separates (100\% TDR at 0.2\% FDR) bonafide and PA iris images by a higher margin compared to the NIR and VIS modalities.

\begin{table}[]	
	\caption{Data distribution among train, validation and test sets for all experiments (intra-attack and cross-attack scenarios). Here, CC is Cosmetic Contacts.}
	\label{table:Experimental-Setup}
	\resizebox{\columnwidth}{!}{%
		\begin{tabular}{|l|l|l|l|l|l|l|}
			\hline
			\multirow{2}{*}{\textbf{Experiments}} & \multicolumn{2}{c|}{\textbf{Train Set}} & \multicolumn{2}{c|}{\textbf{Validation Set}} & \multicolumn{2}{c|}{\textbf{Test Set}} \\ \cline{2-7} 
			& \textbf{Bonafide} & \textbf{PAs} & \textbf{Bonafide} & \textbf{PAs} & \textbf{Bonafide} & \textbf{PAs} \\ \hline
			\begin{tabular}[c]{@{}l@{}}Intra-EXP 1 \\ (Both Artificial Eyes \& CC)\end{tabular} & 404 & 100 & 101 & 25 & 218 & 54 \\ \hline
			\begin{tabular}[c]{@{}l@{}}Intra-EXP 2\\ (Only Artificial Eyes)\end{tabular} & 435 & 35 & 145 & 12 & 146 & 12 \\ \hline
			\begin{tabular}[c]{@{}l@{}}Intra-EXP 3\\ (Only CC)\end{tabular} & 435 & 72 & 145 & 24 & 146 & 24 \\ \hline
			\begin{tabular}[c]{@{}l@{}}Cross-EXP 1\\ (CC are unknown)\end{tabular} & 435 & 41 & 145 & 18 & 146 & 120 \\ \hline
			\begin{tabular}[c]{@{}l@{}}Cross-EXP 2\\ (Artificial eyes are unknown)\end{tabular} & 435 & 84 & 145 & 36 & 146 & 59 \\ \hline
		\end{tabular}
	}
\end{table}

\subsection{Cross-attack Setup and Results}
\begin{figure*}
	\centering
	\includegraphics[width=0.87\linewidth]{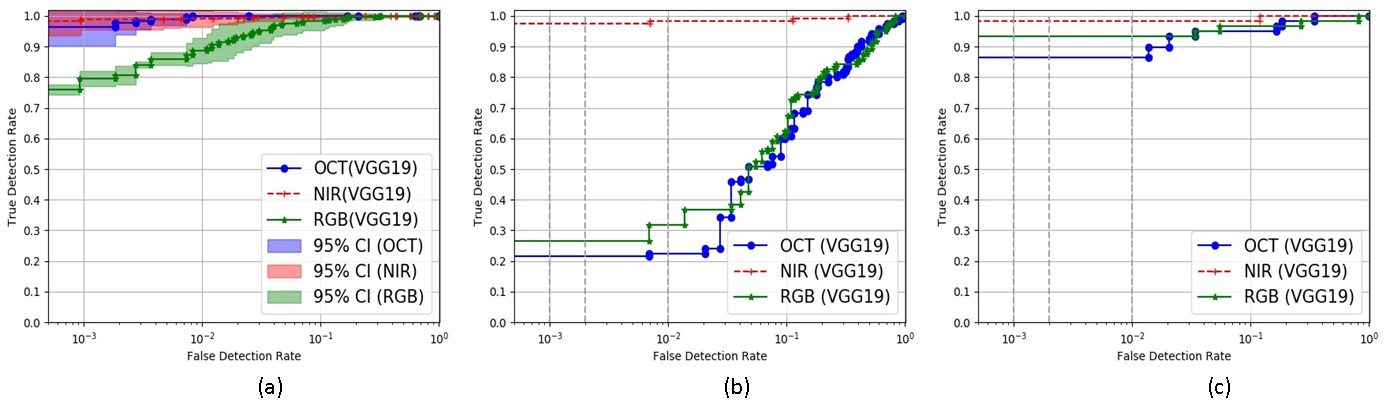}
	\caption{ROC curves of (a) Intra-EXP 1, (b) Cross-EXP 1 and (c) Cross-EXP 2 experiments using \textbf{VGG19} architecture. The first ROC plot (a) also shows the confidence interval of 95\%. NIR imaging is more efficient in discriminating bonafide and PA samples on this network.}
	\label{fig:VGG19-Results}
\end{figure*}

\begin{figure*}
	\centering
	\includegraphics[width=0.87\linewidth]{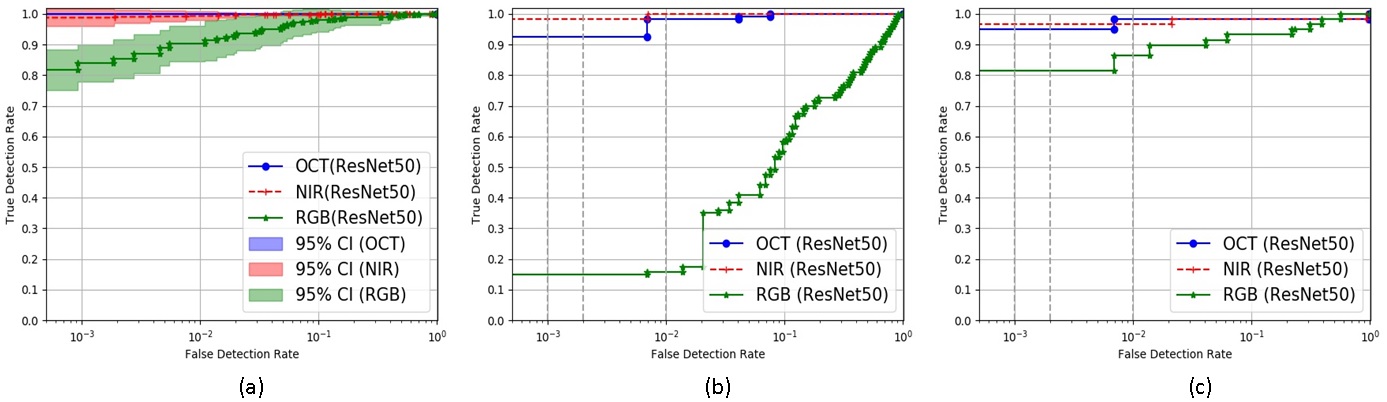}
	\caption{ROC curves of (a) Intra-EXP 1, (b) Cross-EXP 1 and (c) Cross-EXP 2 experiments using \textbf{ResNet50} architecture. OCT imaging results in better performance in distinguishing bonafide and PA images in the intra-attack scenario (a), whereas NIR imaging performs the best in the cross-attack scenario (b and c). }
	\label{fig:ResNet50-Results}
\end{figure*}

\begin{figure*}[h!]
	\centering
	\includegraphics[width=0.87\linewidth]{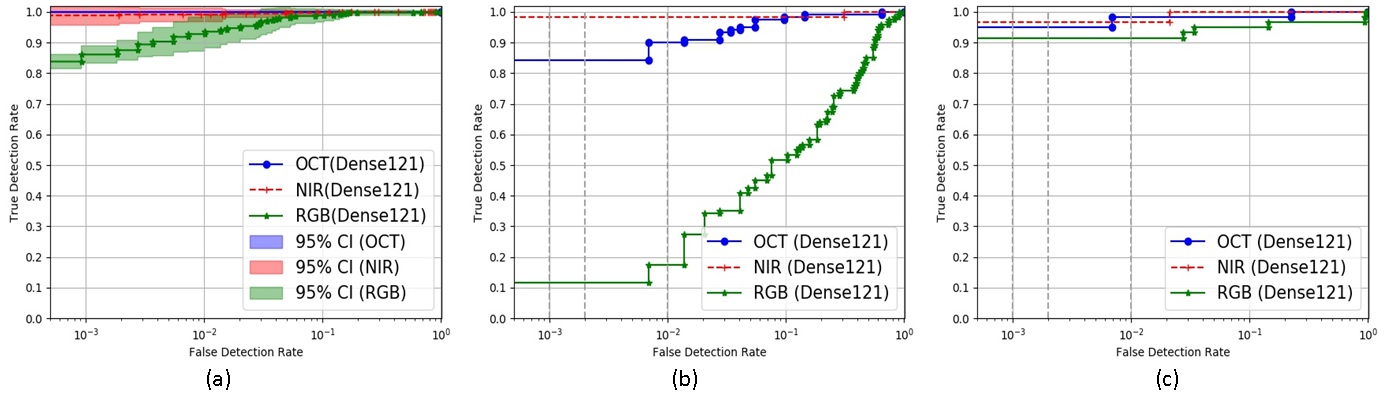}
	\caption{ROC curves of (a) Intra-EXP 1, (b) Cross-EXP 1 and (c) Cross-EXP 2 experiments using \textbf{DenseNet121} architecture. OCT imaging results in better performance in distinguishing bonafide and PA images in the intra-attack scenario (a), whereas NIR imaging performs the best in the cross-attack scenario (b and c).}
	\label{fig:DesNet121-Results}
\end{figure*}

To perform the cross-attack (generalization to unknown attacks) analysis, two experiments are conducted: Cross-EXP 1 and Cross-EXP 2. In the first experiment (Cross-EXP 1), training is performed on bonafide and artificial eye images, and testing is done on bonafide and cosmetic contact lens images. Bonafide images are split in a 60:20:20 ratio for the training, validation and test sets, respectively. Artificial eye images are split in a 70:30 ratio for the training and validation sets, respectively. All cosmetic contact images constitute the test set. In the second experiment (Cross-EXP 2), training is performed on bonafide and cosmetic contact lens images, and testing is done on bonafide and artificial eye images. Bonafide images are split in the same way as Cross-EXP 1. Cosmetic contact lens images are split in a 70:30 proportion for the training and validation sets, respectively. All artificial eye images are used in the test set. Further details of both the experimental setups are given in Table \ref{table:Experimental-Setup}. The TDR at 0.2\% FDR and the ACER for VGG19, ResNet50 and DenseNet121 architectures are provided in Table \ref{table:All-Results}. ROC curves of all three architectures for the two experiments are shown in Figures \ref{fig:VGG19-Results}(b) and \ref{fig:VGG19-Results}(c), \ref{fig:ResNet50-Results}(b) and \ref{fig:ResNet50-Results}(c), and \ref{fig:DesNet121-Results}(b) and \ref{fig:DesNet121-Results}(c), respectively.

In the cross-attack scenario, the best results are observed on NIR images, followed by OCT images and then VIS images. Basically, the OCT and VIS modalities failed in detecting cosmetic contact images when training is performed using artificial eye PAs (see Figures \ref{fig:VGG19-Results}(b), \ref{fig:ResNet50-Results}(b) and \ref{fig:DesNet121-Results}(b)). The feature sub-spaces of bonafide samples and cosmetic contact lens seem to overlap (middle column of Figure \ref{fig:t-sne-plots}). However, when classifiers are trained on cosmetic contact images (Figure \ref{fig:VGG19-Results}(c), \ref{fig:ResNet50-Results}(c) and \ref{fig:DesNet121-Results}(c)), they can detect artificial eye PAs as feature sub-space of artificial eyes seems to be well separated from that of bonafide samples (last column of Figure \ref{fig:t-sne-plots}). Difficulty in detecting cosmetic contact PAs is also reflected in the Intra-EXP 2 and Intra-EXP 3 experiments. ResNet50 and DenseNet121 architectures are better suited for the cross-attack scenario than the VGG19 network, as a higher number of trainable parameters are present in VGG19 and the training data is insufficient. As the networks are pre-trained on the ImageNet dataset (containing VIS images), trainable parameters converge in the case of VIS and NIR images, but fail to converge for OCT images due to the fundamentally different image modality (Figure 5(a)).  

\begin{table*}[]	
	\caption{TDR (\%) at 0.2\% FDR and ACER of all experiments (intra-attack and cross-attack scenarios) when using VGG19, ResNet50 and DenseNet121 architectures.}
	\label{table:All-Results}
	\resizebox{\textwidth}{!}{%
		\begin{tabular}{|l|l|l|l|l|l|l|l|l|l|l|}
			\hline
			&  & \multicolumn{3}{c|}{VGG19} & \multicolumn{3}{c|}{ResNet50} & \multicolumn{3}{c|}{DenseNet121} \\ \cline{3-11} 
			\multirow{-2}{*}{Experiments} & \multirow{-2}{*}{\begin{tabular}[c]{@{}l@{}}Evaluation \\ Measure\end{tabular}} & OCT & NIR & RGB & OCT & NIR & RGB & OCT & NIR & RGB \\ \hline
			& \cellcolor[HTML]{FDE6E6}ACER & \cellcolor[HTML]{FDE6E6}0.08 ± 0.15 & \cellcolor[HTML]{FDE6E6}\textbf{0.02 ± 0.01} & \cellcolor[HTML]{FDE6E6}0.09 ± 0.03 & \cellcolor[HTML]{FDE6E6}\textbf{0.00 ± 0.00} & \cellcolor[HTML]{FDE6E6}0.00 ± 0.01 & \cellcolor[HTML]{FDE6E6}0.08 ± 0.00 & \cellcolor[HTML]{FDE6E6}0.02 ± 0.03 & \cellcolor[HTML]{FDE6E6}\textbf{0.02 ± 0.02} & \cellcolor[HTML]{FDE6E6}0.07 ± 0.02 \\ \cline{2-11} 
			\multirow{-2}{*}{\begin{tabular}[c]{@{}l@{}}Intra-EXP 1 \\ (Both Artificial \& CC)\end{tabular}} & TDR & \textbf{100 ± 0.00} & 97.99 ± 2.66 & 82.58 ± 6.88 & \textbf{100 ± 0.00} & 97.33 ± 3.88 & 89.62 ± 3.62 & \textbf{100 ± 0.00} & 97.66 ± 3.26 & 86.66 ± 3.59 \\ \hline
			& \cellcolor[HTML]{FDE6E6}ACER & \cellcolor[HTML]{FDE6E6}\textbf{0.00} & \cellcolor[HTML]{FDE6E6}\textbf{0.00} & \cellcolor[HTML]{FDE6E6}\textbf{0.00} & \cellcolor[HTML]{FDE6E6}\textbf{0.00} & \cellcolor[HTML]{FDE6E6}\textbf{0.00} & \cellcolor[HTML]{FDE6E6}0.04 & \cellcolor[HTML]{FDE6E6}\textbf{0.00} & \cellcolor[HTML]{FDE6E6}\textbf{0.00} & \cellcolor[HTML]{FDE6E6}\textbf{0.00} \\ \cline{2-11} 
			\multirow{-2}{*}{\begin{tabular}[c]{@{}l@{}}Intra-EXP 2 \\ (Only Artificial Eyes)\end{tabular}} & TDR & \textbf{100} & \textbf{100} & \textbf{100} & \textbf{100} & \textbf{100} & \textbf{100} & \textbf{100} & \textbf{100} & \textbf{100} \\ \hline
			& \cellcolor[HTML]{FDE6E6}ACER & \cellcolor[HTML]{FDE6E6}\textbf{0.00} & \cellcolor[HTML]{FDE6E6}\textbf{0.00} & \cellcolor[HTML]{FDE6E6}0.03 & \cellcolor[HTML]{FDE6E6}\textbf{0.00} & \cellcolor[HTML]{FDE6E6}\textbf{0.00} & \cellcolor[HTML]{FDE6E6}\textbf{0.00} & \cellcolor[HTML]{FDE6E6}\textbf{0.00} & \cellcolor[HTML]{FDE6E6}\textbf{0.00} & \cellcolor[HTML]{FDE6E6}0.03 \\ \cline{2-11} 
			\multirow{-2}{*}{\begin{tabular}[c]{@{}l@{}}Intra-EXP 3 \\ (Only CC)\end{tabular}} & TDR & \textbf{100} & \textbf{100} & 95.83 & \textbf{100} & \textbf{100} & \textbf{100} & \textbf{100} & \textbf{100} & \textbf{100} \\ \hline
			& \cellcolor[HTML]{FDE6E6}ACER & \cellcolor[HTML]{FDE6E6}0.39 & \cellcolor[HTML]{FDE6E6}\textbf{0.01} & \cellcolor[HTML]{FDE6E6}0.19 & \cellcolor[HTML]{FDE6E6}0.20 & \cellcolor[HTML]{FDE6E6}\textbf{0.01} & \cellcolor[HTML]{FDE6E6}0.27 & \cellcolor[HTML]{FDE6E6}0.16 & \cellcolor[HTML]{FDE6E6}\textbf{0.01} & \cellcolor[HTML]{FDE6E6}0.30 \\ \cline{2-11} 
			\multirow{-2}{*}{\begin{tabular}[c]{@{}l@{}}Cross-EXP 1 \\ (CC are unknown)\end{tabular}} & TDR & 21.66 & \textbf{97.58} & 26.66 & 92.50 & \textbf{98.38} & 15.00 & 84.16 & \textbf{98.38} & 11.66 \\ \hline
			& \cellcolor[HTML]{FDE6E6}ACER & \cellcolor[HTML]{FDE6E6}0.06 & \cellcolor[HTML]{FDE6E6}\textbf{0.03} & \cellcolor[HTML]{FDE6E6}0.04 & \cellcolor[HTML]{FDE6E6}\textbf{0.01} & \cellcolor[HTML]{FDE6E6}0.02 & \cellcolor[HTML]{FDE6E6}0.07 & \cellcolor[HTML]{FDE6E6}0.05 & \cellcolor[HTML]{FDE6E6}\textbf{0.01} & \cellcolor[HTML]{FDE6E6}0.04 \\ \cline{2-11} 
			\multirow{-2}{*}{\begin{tabular}[c]{@{}l@{}}Cross-EXP 2\\ (Artificial eyes are unknown)\end{tabular}} & TDR & 86.44 & \textbf{98.38} & 93.22 & 94.91 & \textbf{96.77} & 81.35 & 94.91 & \textbf{96.77} & 91.52 \\ \hline
		\end{tabular}
	}
\end{table*}

The main findings of the comparative analysis are:
\begin{enumerate}
	\item In the intra-attack scenario, when PAs are known and used during training, OCT images provide more discriminative information for distinguishing between bonafide and PA samples. However, NIR imaging provides better generalizability across unknown iris PA attacks.
	\item Cosmetic contact PAs are difficult to detect compared to artificial eyes, especially on VIS images.
	\item ResNet50 and DenseNet121 architectures are well-suited for iris PA detection in the OCT imaging modality possibly due to the smaller number of trainable parameters compared to VGG-19.% (OCT, NIR and VIS) and in both the scenarios (intra-attack and cross-attack).
\end{enumerate}

%In brief, all three architectures perfectly classify the bonafide images from the presentation attacks images on the OCT data under the intra-attack scenario. However, in cross-attack scenario (unseen attack), NIR images provide more discriminative information. 

\section{CNN Visualization}
\begin{figure*}
	\centering
	\includegraphics[scale = 0.7]{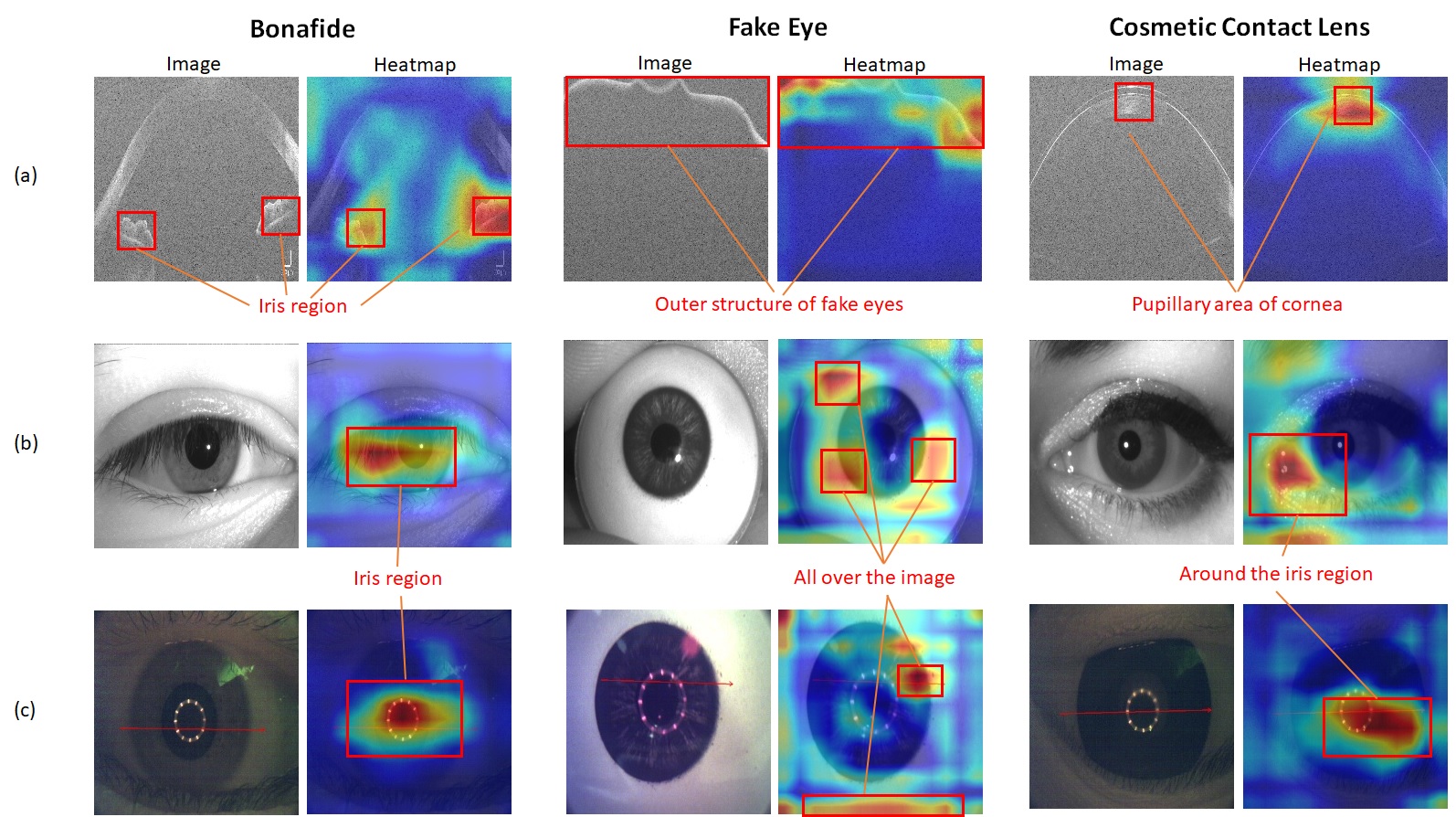}
	\caption{(a) OCT, (b) NIR and (c) VIS images and their corresponding fixation regions for bonafide, artificial eyes and cosmetic contact lens samples. Red in the heatmaps represents high priority (high CNN activations) regions considered by the CNN architecture. Blue represents low priority regions. Red boxes mark the high priority regions. Different regions of focus help the CNN architecture to differentiate between bonafide and PA iris images.}
	\label{fig:Grad-cam}
\end{figure*}

The performance of all three architectures is nearly perfect on OCT and NIR images. To further analyze the results, we generate heatmaps \cite{Selvaraju2017} and t-SNE plots \cite{Maaten2008}. Heatmaps provide the salient regions in OCT, NIR and VIS images where the classifier (ResNet50) focused on, in order to discriminate PAs from bonafide samples. Heatmaps are generated using Grad-CAM \cite{Selvaraju2017}. Grad-CAM uses a gradient of the loss function and backpropagates it through the convolutional layers to generate activations on the input image. OCT, NIR and VIS images of a bonafide, artificial eye and cosmetic contact lens are shown along with their heatmaps in Figure \ref{fig:Grad-cam}. 
%Red color gradient represents the regions of high CNN activations, whereas blue color gradient shows regions of low CNN activations. 
In the case of OCT images (Figure \ref{fig:Grad-cam}(a)), the heatmap of the bonafide image highlights the iris regions, which is the most discriminative region compared to OCT PA images. The heatmap of an artificial eye image focuses over the outer structure. Cosmetic contact lens conceals the underlying iris pattern (partially or fully), which causes the focus to shift over to the corneal region corresponding to the pupil. In the case of NIR and VIS imaging (Figure \ref{fig:Grad-cam}(a) and \ref{fig:Grad-cam}(b)), heatmaps of bonafide sample focus over the iris pattern. For an artificial eye image, the heatmap is activated all over the image, whereas for a textured contact lens more emphasis is given to the circumference of the iris. Different regions of focus for different categories (bonafide and PA) aid the CNN architecture to discriminate between them. %Heatmap of OCT images shows disparate regions of focus in bonafide and PA images resulting in better performance in the intra-attack scenario.

\begin{figure*}
	\centering
	\includegraphics[scale = 0.5]{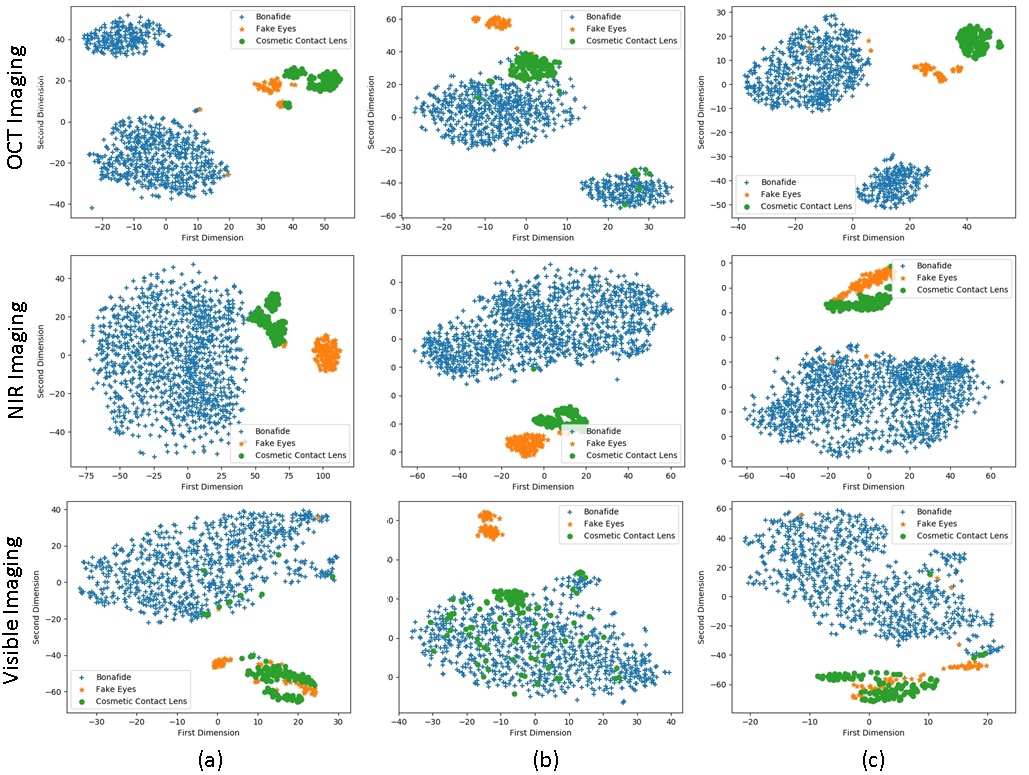}
	\caption{t-SNE plots of Intra-EXP 1, Cross-EXP 1 and Cross-EXP 2 test data pertaining to OCT, NIR and VIS imaging. 2048 dimensions of features from the average pooling layer (penultimate layer) of ResNet50 network are reduced to two dimensions for visualization. Features of bonafide and PAs from OCT images are well separated in Intra-EXP 1 and Cross-EXP 2 experiments. NIR images show good separation in all three experiments. Features from VIS images are overlapping between the bonafide and PA categories (especially in the Cross-EXP 2 experiment). More the separation of features, better the classification.}
	\label{fig:t-sne-plots}
\end{figure*}

After visualizing activations on the input image, we also visualize the CNN features using a t-SNE plot \cite{Maaten2008}. The CNN features are extracted from the average pooling layer (penultimate  layer, a layer before the last fully connected layer) of the ResNet50 architecture. The dimensionality of the features is 2048, which is reduced to two dimensions using t-Distributed Stochastic Neighbor Embedding (t-SNE). The t-SNE plots are shown in Figure \ref{fig:t-sne-plots}. These t-SNE plots correspond to Intra-EXP 1 (first column), Cross-EXP 1 (second column) and Cross-EXP 2 (third column) test data. Distribution of bonafide, artificial eyes and cosmetic contact images are observed to be well separated in OCT imaging in the case of Intra-EXP 1 and Cross-EXP 2 experiments. Separation of these features is also prominent in NIR imaging under the cross-attack scenario (Cross-EXP 1 and Cross-EXP 2). Features in the case of Cross-EXP 1 experiment overlap for VIS images. These plots substantiate our observations that OCT imaging works efficiently in the intra-attack scenario and moderately in the cross-attack scenario, while NIR imaging generalizes well in the cross-attack scenario. 

\section{Conclusion and Future Work}
In this paper, we described the use of the OCT imaging modality for iris PA detection. By comparative analysis against other imaging modalities (traditional NIR and VIS), we determined that OCT is a viable solution for iris PA detection. Extensive experiments were conducted both in the intra-attack and cross-attack scenarios using three state-of-the-art deep architectures, and results were analyzed using CNN visualizations (heatmaps and t-SNE plots). Future work will involve collecting OCT data from more subjects and other types of PAs. Hardware cost continues to be a barrier for the use of OCT in iris recognition applications. However, as sophisticated presentation attacks are launched in the future, the OCT modality is likely to be of great benefit.  
%VIS imaging failed to provide useful information for differentiating bonafides from PAs, especially in the cross-attack scenario. 

\section*{Acknowledgment}
This research is based upon work supported in part by the Office of the Director of National Intelligence (ODNI), Intelligence Advanced Research Projects Activity (IARPA), via IARPA R\&D Contract No. 2017 - 17020200004. The IRB number is IRB00123321. The views and conclusions contained herein are those of the authors and should not be interpreted as necessarily representing the official policies, either expressed or implied, of ODNI, IARPA, or the U.S. Government. The U.S. Government is authorized to reproduce and distribute reprints for governmental purposes notwithstanding any copyright annotation therein.

\balance
{\small
	\bibliographystyle{ieee}
	\bibliography{OCT-NIR-RGB-Analysis-V10}

\begin{thebibliography}{10}\itemsep=-1pt

\bibitem{Odin}
{Intelligence Advanced Research Projects Activity (IARPA), ODNI:
  IARPA-BAA-16-04 (Thor)}.
\newblock https://www.iarpa.gov/index.php/research-programs/odin/odin-baa.

\bibitem{OCTsensor}
{THORLabs Telesto series (TEL1325LV2) Spectral domain {OCT} scanner.}
\newblock
  https://www.thorlabs.com/catalogpages/Obsolete/2017/TEL1325LV2-BU.pdf.

\bibitem{Chen2018a}
C.~Chen and A.~Ross.
\newblock A multi-task convolutional neural network for joint iris detection
  and presentation attack detection.
\newblock {\em Winter Applications of Computer Vision Workshops (WACV-W)},
  pages 44--51, 2018.

\bibitem{Chen2012}
R.~Chen, X.~Lin, and T.~Ding.
\newblock Liveness detection for iris recognition using multispectral images.
\newblock {\em Pattern Recognition Letters (PRL)}, 33(12):1513--1519, 2012.

\bibitem{Connell2013}
J.~{Connell}, N.~{Ratha}, J.~{Gentile}, and R.~{Bolle}.
\newblock Fake iris detection using structured light.
\newblock {\em International Conference on Acoustics, Speech and Signal
  Processing (ICASSP)}, pages 8692--8696, 2013.

\bibitem{Czajka2019}
A.~Czajka and B.~Becker.
\newblock {\em Application of Dynamic Features of the Pupil for Iris
  Presentation Attack Detection}, pages 151--168.
\newblock Springer International Publishing, 2019.

\bibitem{Czajka2018}
A.~Czajka and K.~W. Bowyer.
\newblock Presentation attack detection for iris recognition: An assessment of
  the state of the art.
\newblock {\em {ACM} Computing Surveys}, 1(1):1--35, 2018.

\bibitem{LivDet2020}
P.~{Das}, J.~{McGrath}, A.~B. Z.~{Fang}, G.~{Jang}, A.~{Mohammadi},
  S.~{Purnapatra}, D.~{Yambay}, S.~{Marcel}, M.~{Trokielewicz},
  P.~{Maciejewicz}, K.~{Bowyer}, A.~{Czajka}, S.~{Schuckers}, J.~{Tapia},
  S.~{Gonzalez}, M.~{Fang}, N.~{Damer}, F.~{Boutros}, A.~{Kuijper},
  R.~{Sharma}, C.~{Chen}, and A.~{Ross}.
\newblock {Iris Liveness Detection Competition (LivDet-Iris) -- The 2020
  Edition}.
\newblock {\em International Joint Conference on Biometrics (IJCB)}, 2020.

\bibitem{Gragnaniello2015b}
D.~Gragnaniello, C.~Sansone, and L.~Verdoliva.
\newblock Iris liveness detection for mobile devices based on local
  descriptors.
\newblock {\em Pattern Recognition Letters (PRL)}, 57:81--87, 2015.

\bibitem{He2015}
K.~He, X.~Zhang, S.~Ren, and J.~Sun.
\newblock Deep residual learning for image recognition.
\newblock {\em Conference on Computer Vision and Pattern Recognition (CVPR)},
  pages 770--778, 2016.

\bibitem{Hoffman2019}
S.~Hoffman, R.~Sharma, and A.~Ross.
\newblock Iris + ocular: Generalized iris presentation attack detection using
  multiple convolutional neural networks.
\newblock {\em International Conference on Biometrics (ICB)}, 2019.

\bibitem{Hsieh2018}
S.-H. Hsieh, Y.~Li, W.~Wang, and C.-H. Tien.
\newblock A novel anti-spoofing solution for iris recognition toward cosmetic
  contact lens attack using spectral {ICA} analysis.
\newblock {\em Sensors}, 18:795--810, 2018.

\bibitem{Huang1991}
D.~Huang, E.~A. Swanson, C.~P. Lin, J.~S. Schuman, W.~G. Stinson, W.~Chang,
  M.~R. Hee, T.~Flotte, K.~Gregory, C.~A. Puliafito, and et~al.
\newblock Optical coherence tomography.
\newblock {\em Science}, 254:1178--1181, 1991.

\bibitem{Huang2018}
G.~{Huang}, Z.~{Liu}, L.~v.~d. {Maaten}, and K.~Q. {Weinberger}.
\newblock Densely connected convolutional networks.
\newblock {\em Conference on Computer Vision and Pattern Recognition (CVPR)},
  pages 2261--2269, 2017.

\bibitem{Hughes2013}
K.~{Hughes} and K.~W. {Bowyer}.
\newblock Detection of contact-lens-based iris biometric spoofs using stereo
  imaging.
\newblock {\em Hawaii International Conference on System Sciences (HICSS)},
  pages 1763--1772, 2013.

\bibitem{Kanematsu2007}
M.~Kanematsu, H.~Takano, and K.~Nakamura.
\newblock Highly reliable liveness detection method for iris recognition.
\newblock {\em SICE Annual Conference}, pages 361--364, 2007.

\bibitem{Komogortsev2013}
O.~Komogortsev and A.~Karpov.
\newblock Liveness detection via oculomotor plant characteristics: Attack of
  mechanical replicas.
\newblock {\em International Conference on Biometrics (ICB)}, pages 1--8, 2013.

\bibitem{Lee2006}
S.~J. Lee, K.~R. Park, and J.~Kim.
\newblock Robust fake iris detection based on variation of the reflectance
  ratio between the iris and the sclera.
\newblock {\em Biometrics Symposium: Special Session on Research at the
  Biometric Consortium Conference}, pages 1--6, 2006.

\bibitem{Menotti2015}
D.~Menotti, G.~Chiachia, A.~da~Silva~Pinto, W.~R. Schwartz, H.~Pedrini, A.~X.
  Falcao, and A.~Rocha.
\newblock Deep representations for iris, face, and fingerprint spoofing
  detection.
\newblock {\em Transactions on Information Forensics and Security (TIFS)},
  10:864--879, 2015.

\bibitem{Moolla2019}
Y.~Moolla, L.~Darlow, A.~Sharma, A.~Singh, and J.~V.~D. Merwe.
\newblock Optical coherence tomography for fingerprint presentation attack
  detection.
\newblock {\em Handbook of Biometric Anti-Spoofing}, pages 49--70, 2019.

\bibitem{Park2007}
J.~H. Park and M.-G. Kang.
\newblock Multispectral iris authentication system against counterfeit attack
  using gradient-based image fusion.
\newblock {\em Optical Engineering}, 46(11):1--14, 2007.

\bibitem{Raghavendra2015}
R.~Raghavendra and C.~Busch.
\newblock Robust scheme for iris presentation attack detection using multiscale
  binarized statistical image features.
\newblock {\em Transactions on Information Forensics and Security (TIFS)},
  10:703--715, 2015.

\bibitem{Raja2015}
K.~B. {Raja}, R.~{Raghavendra}, and C.~{Busch}.
\newblock Video presentation attack detection in visible spectrum iris
  recognition using magnified phase information.
\newblock {\em Transactions on Information Forensics and Security (TIFS)},
  10(10):2048--2056, 2015.

\bibitem{Ramachandra2014}
R.~Ramachandra and C.~Busch.
\newblock Presentation attack detection on visible spectrum iris recognition by
  exploring inherent characteristics of light field camera.
\newblock {\em International Joint Conference on Biometrics (IJCB)}, 2014.

\bibitem{Ross2019}
A.~Ross, S.~Banerjee, C.~Chen, A.~Chowdhury, V.~Mirjalili, R.~Sharma,
  T.~Swearingen, and S.~Yadav.
\newblock Some research problems in biometrics: The future beckons.
\newblock {\em International Conference on Biometrics (ICB)}, 2019.

\bibitem{Selvaraju2017}
R.~R. {Selvaraju}, M.~{Cogswell}, A.~{Das}, R.~{Vedantam}, D.~{Parikh}, and
  D.~{Batra}.
\newblock {Grad-CAM}: Visual explanations from deep networks via gradient-based
  localization.
\newblock {\em International Conference on Computer Vision (ICCV)}, pages
  618--626, 2017.

\bibitem{Sequeira2016}
A.~F. {Sequeira}, S.~{Thavalengal}, J.~{Ferryman}, P.~{Corcoran}, and J.~S.
  {Cardoso}.
\newblock A realistic evaluation of iris presentation attack detection.
\newblock {\em Conference on Telecommunications and Signal Processing (TSP)},
  pages 660--664, 2016.

\bibitem{Sharma2020}
R.~Sharma and A.~Ross.
\newblock {D-NetPAD}: An explainable and interpretable iris presentation attack
  detector.
\newblock {\em International Joint Conference on Biometrics (IJCB)}, 2020.

\bibitem{Simonyan2015}
K.~Simonyan and A.~Zisserman.
\newblock Very deep convolutional networks for large{-}scale image recognition.
\newblock {\em International Conference on Learning Representations (ICLR)},
  2015.

\bibitem{Song2019}
G.~Song, K.~K. Chu, S.~Kim, M.~Crose, B.~Cox, E.~T. Jelly, N.~Ulrich, and
  A.~Wax.
\newblock First clinical application of low-cost {OCT}.
\newblock {\em Translational vision science and technology (TVST)}, 8(3):61,
  2019.

\bibitem{Shejin2016}
S.~Thavalengal, T.~Nedelcu, P.~Bigioi, and P.~Corcoran.
\newblock Iris liveness detection for next generation smartphones.
\newblock {\em Transactions on Consumer Electronics (TCE)}, 62:95--102, 2016.

\bibitem{Maaten2008}
L.~van~der Maaten and G.~Hinton.
\newblock Visualizing high-dimensional data using t-sne.
\newblock {\em Journal of Machine Learning Research (JMLR)}, pages 2579--2605,
  2008.

\bibitem{Vyas2019}
R.~Vyas, T.~Kanumuri, and G.~Sheoran.
\newblock Cross spectral iris recognition for surveillance based applications.
\newblock {\em Multimedia Tools and Applications (MTA)}, 78(5):5681--5699,
  2019.

\bibitem{LivDet2017}
D.~{Yambay}, B.~{Becker}, N.~{Kohli}, D.~{Yadav}, A.~{Czajka}, K.~W. {Bowyer},
  S.~{Schuckers}, R.~{Singh}, M.~{Vatsa}, A.~{Noore}, D.~{Gragnaniello},
  C.~{Sansone}, L.~{Verdoliva}, L.~{He}, Y.~{Ru}, H.~{Li}, N.~{Liu}, Z.~{Sun},
  and T.~{Tan}.
\newblock {LivDet iris 2017} — iris liveness detection competition 2017.
\newblock {\em International Joint Conference on Biometrics (IJCB)}, pages
  733--741, 2017.

\bibitem{Zhang2015}
H.~Zhang, Z.~Sun, T.~Tan, and J.~Wang.
\newblock Learning hierarchical visual codebook for iris liveness detection.
\newblock {\em International Joint Conference on Biometrics (IJCB)}, pages
  1--8, 2015.

\end{thebibliography}
}

\end{document}